\documentclass[lettersize,journal]{IEEEtran}  

\usepackage[english]{babel}
\usepackage{algorithmic}
\usepackage{graphicx} 
\graphicspath{{ {./images/} }}
\usepackage{epsfig} 
\usepackage{hyperref}
\usepackage{amsmath} 
\usepackage{amssymb}  
\usepackage{bm}
\usepackage{color}
\usepackage{subcaption}
\usepackage{caption}
\usepackage{cite}
\usepackage{xfrac}
\usepackage[font=footnotesize,labelfont=footnotesize]{caption}

\usepackage{algorithm}
\usepackage{algorithmic}
\usepackage{array}
\usepackage{makecell}

\newcolumntype{C}[1]{>{\centering\arraybackslash}p{#1}}
\vspace{-0.5em}

\usepackage{soul}
\makeatletter 
\newcount\SOUL@minus
\makeatother  

\usepackage{textcomp}
\def\BibTeX{{\rm B\kern-.05em{\sc i\kern-.025em b}\kern-.08em
    T\kern-.1667em\lower.7ex\hbox{E}\kern-.125emX}}

\begin{document}
\title{Deep Robust Koopman Learning from Noisy Data
}
\author{Aditya~Singh,~Rajpal~Singh,~\IEEEmembership{Graduate Student Member,~IEEE}~and~Jishnu~Keshavan,~\IEEEmembership{Member,~IEEE}
\thanks{This research was supported in part by the Science and Engineering Research Board (SERB) Core Research Grant.}
 \thanks{The authors are with the Department of Mechanical Engineering, Indian Institute of Science, Bangalore, Karnataka~560012, India (email : adityasingh2@iisc.ac.in,
 rajpalsingh@iisc.ac.in, kjishnu@iisc.ac.in).}
}

\maketitle
\begin{abstract}
Koopman operator theory has emerged as a leading data-driven approach that relies on a judicious choice of observable functions to realize global linear representations of nonlinear systems in the lifted observable space. However, real-world data is often noisy, making it difficult to obtain an accurate and unbiased approximation of the Koopman operator. The Koopman operator generated from noisy datasets is typically corrupted by noise-induced bias that severely degrades prediction and downstream tracking performance. In order to address this drawback, this paper proposes a novel autoencoder-based neural architecture to jointly learn the appropriate lifting functions and the reduced-bias Koopman operator from noisy data\color{black}. The architecture initially learns the Koopman basis functions that are consistent for both the forward and backward temporal dynamics of the system. Subsequently, by utilizing the learned forward and backward temporal dynamics, the Koopman operator is synthesized with a reduced bias making the method more robust to noise compared to existing techniques. Theoretical analysis is used to demonstrate  significant bias reduction in the presence of training noise. Dynamics prediction and tracking control simulations are conducted for multiple serial manipulator arms, including performance comparisons with leading alternative designs, to demonstrate its robustness under various noise levels. Experimental studies with the Franka FR3 7-DoF manipulator arm are further used to demonstrate the effectiveness of the proposed approach in a practical setting. 

\end{abstract}


\IEEEpeerreviewmaketitle

\section{Introduction}
\IEEEPARstart{C}{ontrol} of nonlinear dynamical systems remains a persistent challenge across various scientific and engineering fields. Despite extensive research, a comprehensive and generalized mathematical framework for managing such complex systems has yet to be established. Linear control theory provides powerful tools for analysis and control of linear systems, thus motivating efforts to develop approaches that cast nonlinear dynamics within a linear framework. Among these approaches, Koopman operator theory~\cite{Koopman1931hamiltonian} has gained prominence for its ability to derive global linear representations of nonlinear systems. Unlike local linearization techniques, such as Taylor series approximations, Koopman operator theory encodes system dynamics linearly across the entire state space, delivering markedly improved predictive performance~\cite{korda2018linear, book}.

Koopman theory employs a set of nonlinear basis functions that evolve linearly in the transformed observable space under the action of the Koopman operator~{\cite{Koopman1931hamiltonian}}. However, a complete representation of system dynamics in the observable space often demands an infinite set of observable functions, which is impractical for real-world applications. To overcome this limitation, several data-driven methods such as Dynamic Mode Decomposition (DMD){\cite{dmdtu2013}}, Dynamic Mode Decomposition with Control (DMDc){\cite{dmdcproctor2016}}, Extended Dynamic Mode Decomposition (EDMD)\cite{edmdwilliams2014}, and Sparse Identification of Nonlinear Dynamical System (SINDy)\cite{sindybrunton2016}, \color{black}have been developed to construct finite-dimensional approximations of Koopman linear models. However, the aforementioned data-driven methods often rely on manual choices to find the required finite set of observable functions. While prior knowledge of simpler systems can guide such manual selection, identification of an optimal set becomes increasingly challenging and tedious for complex systems. To address this drawback, various learning-based approaches have been introduced for selection of an optimal set of observable functions~\cite{sah2025overview}. For example, dictionary learning has been applied to implement EDMD in \cite{li2017extended}, while autoencoders have been used for SINDy in~\cite{champion2019data} and Koopman eigenvalue decomposition in~\cite{lusch2018deep}. The efficacy of these learning frameworks in generating high-fidelity linear Koopman models has fostered their application in the modeling and control of robotic systems such as manipulators~\cite{sah2024real}, quadrotors~\cite{folkestadEpisodic, folkestad2022koopnet}, and soft robots~\cite{bruder2019nonlinear, haggerty2020modeling, castano2020control}.


An important consideration in employing data-driven frameworks for accurate learning of the Koopman operator is their critical dependency on the nature of the underlying data. In particular, it is well-known that datasets acquired from real-world systems are often noisy which leads to inaccurate predictions of the Koopman operator \cite{Dawson2014CharacterizingAC}. The noise inherent in the system’s original state space is typically assumed to be zero-mean Gaussian. However, after transformation through nonlinear lifting functions, its statistical properties become ambiguous, deviating from the underlying Gaussian distribution and acquiring a non-zero mean. As a result, the standard assumptions of Gaussian and zero-mean noise no longer hold, leaving the characterization of uncertainty in the lifted state space an open challenge for the research community \cite{Shi2024KoopmanOI}. De-noising the data becomes challenging when the underlying model is unknown \cite{sedehi-2025}, as such approaches typically assume access to ground truth which often does not hold for real-world datasets. Consequently, de-noising may distort the underlying system dynamics \cite{Widmann2015DigitalFD},\cite{doi:10.1177/23312165231192304}. This highlights the need for designing Koopman architectures that are inherently robust to noise, ensuring reliable predictive performance in real-world scenarios. Hence, several Koopman architectures have been proposed recently to deal with noisy data. In particular, the effect of noise on DMD is considered in~\cite{Dawson2014CharacterizingAC}, and several methods including forward-backward DMD for correcting noise-induced bias are subsequently proposed in the same study. However, as DMD only utilizes the physical states of the systems as observables, the zero-mean Gaussian noise model remains valid but makes it incapable of capturing dynamics of highly nonlinear systems. Thus, an extension to this architecture is proposed in forward-backward EDMD with inputs in \cite{lortie2024forwardbackwardextendeddmdasymptotic} although it continues to assume the zero-mean noise in the lifted state. Nonetheless, as mentioned earlier, finding the suitable Koopman observable functions remains a challenge for nonlinear systems without using deep learning architectures. To overcome this drawback, the Deep Koopman Learning using Noisy Data (DKND) scheme is proposed in~\cite{Hao2024DeepKL}, which tunes the basis functions to reduce the effect of noise. Yet, the method overlooks the substantial bias that arises when the Koopman operator is estimated via linear regression under noisy observations. As a result, the learned operator retains considerable bias, which leads to reduced prediction accuracy. Another deep learning-based architecture is proposed in \cite{Sakib2024LearningNS} to minimize a recurrent loss by increasing the roll-out duration, which relies on progressively increasing the time horizon for computing the prediction loss. However, this study neither examines the effects of noise nor offers mathematical guarantees of robustness, and incurs a significant computational burden. An adaptive Koopman framework is proposed in~\cite{singh2024adaptive}, where the nominal Koopman model is modified using online learning to adapt to uncertainties present in measurements and system dynamics. However, this study learns the nominal model offline using noise-less data acquired from simulations, without providing any explicit guarantees for robustness against noise. 
Finally, the study in \cite{pmlr-v119-azencot20a} utilizes an autoencoder to generate consistent Koopman observables by leveraging both forward and backward dynamics of an autonomous system. In contrast to the present study, this formulation does not examine the influence of noise, and therefore offers no theoretical guarantees regarding the noise-robustness properties of the resulting Koopman operator. Further, the approach is confined to autonomous systems and does not extend to control-affine systems. 

To overcome these challenges, we introduce the Deep Robust Koopman Network (DRKN) framework that mitigates the bias induced by noise present in training data. This approach builds on the forward-backward DMD concept from \cite{Dawson2014CharacterizingAC}, and extends the analysis to control-affine systems with nonlinear Koopman observables. By embedding this concept within an autoencoder-based neural network architecture, DRKN enables data-driven discovery of optimal basis functions, while simultaneously learning the Koopman operator corresponding to both the forward and backward temporal dynamics of control-affine systems. The proposed method exploits the analytical properties of the forward–backward dynamical system to realize a more accurate (less biased) estimation of the Koopman operator. In turn, this enhances robustness to noisy data without requiring prior knowledge of the statistical properties of noise in the lifted space, thereby enabling the identification of reliable models for highly nonlinear systems. To the best of the authors’ knowledge, this is the first study to explicitly characterize noise-induced bias in the Koopman operator learnt using nonlinear basis functions within the lifted state space, where the noise no longer remains Gaussian with zero mean. Theoretical analysis is undertaken to provide robustness guarantees with the proposed approach under mild assumptions on the statistical properties of observation noise. Comprehensive simulation and experimental studies, including comparisons with leading alternative designs, are used to demonstrate the efficacy and superior performance of the proposed scheme. 

Thus, the main contributions of our paper are as follows:
\begin{enumerate}
    \item {We propose a novel neural network architecture that employs forward-backward dynamics propagation to learn accurate Koopman operators from noisy data for highly nonlinear control-affine systems.}
    
    \item The proposed architecture employs an autoencoder to automatically learn an optimal basis function set, in contrast to previous forward-backward Koopman frameworks that rely on manual selection of basis functions.
    
    \item {Theoretical analysis of noise robustness is undertaken to explicitly demonstrate bias reduction in the synthesis of the Koopman operator using the proposed scheme.}
    

     \item {The efficacy of the proposed scheme is illustrated through closed-loop trajectory tracking control of several serial manipulator systems. In particular, the scalability of the proposed approach, which is a significant advantage relative to the state-of-the-art, is clearly demonstrated through tracking control experiments on the Franka FR3 7-DoF system.}
\end{enumerate}

The rest of the paper is organised as follows. We begin by introducing Koopman operator theory in Section II. Section III outlines the proposed  Deep Robust Koopman Network (DRKN) architecture. Theoretical analysis, including key assumptions and supporting proofs, is provided in Section IV. Section V provides simulations studies undertaken to demonstrate the noise robustness of the architecture. Section VI showcases the real-world experiments conducted with proposed architecture. Ultimately, conclusions are presented under Section VII.

\section{Preliminaries}
\subsection{Background}
To lay the groundwork, we first provide a concise introduction to the  Koopman operator theory~\cite{Koopman1931hamiltonian}. Consider a nonlinear autonomous system described as:
\begin{eqnarray}
    \label{eq:base_non_linear_sytem}
    \boldsymbol{\dot{x}} = \boldsymbol{f}(\boldsymbol{x}),
\end{eqnarray}
where the state of the system, $\boldsymbol{x} \in \mathbb{X} \subset \mathbb{R}^{n}$ and $\boldsymbol{f}:\mathbb{X} \to \mathbb{X}$ are assumed to be Lipschitz continuous on $\mathbb{X}$. The corresponding discrete dynamics can be given as, $\boldsymbol{{x}}_{k+1} = \boldsymbol{F}(\boldsymbol{x}_k)$, where $\boldsymbol{x}_k$ describes the state at $k^{th}$ time step and $\boldsymbol{S}:\mathbb{X} \to \mathbb{X}$ represents the flow map. 

The evolution of a set of observable functions $\sigma: \mathbb{X} \to \mathbb{C}$, residing within a Hilbert space, is described by the continuous Koopman operator, $\mathcal{K}: \mathbb{C} \to \mathbb{C}$, in a linear manner such that
\begin{eqnarray}    
    \mathcal{K}\circ \sigma{(\boldsymbol{x}_k)} =  \sigma\circ \boldsymbol{S}(\boldsymbol{x}_k) = \sigma(\boldsymbol{x}_{k+1}),  \nonumber
\end{eqnarray}
where $\circ$ represents functional composition. Note that the evolution of the observables in Koopman space is linear. However, the exact Koopman representation requires to be infinite-dimensional. As such, often finite-dimensional approximations of the Koopman operator are used for real world applications.

The same theoretical concepts can also be extended to a system with control inputs. Consider a control-affine system of the form:
\begin{eqnarray}
        \label{eq:base_affine}
    \boldsymbol{\dot{x}} = \boldsymbol{f}_0(\boldsymbol{x}) + \sum_{i=1}^{m}\boldsymbol{f}_i(\boldsymbol{x}){u}_{i},
\end{eqnarray}
where $\boldsymbol{u} = [u_1, u_2, ..., u_m]^{\top} \in \mathbb{R}^m$ represents the control input to the system, and $\boldsymbol{f}_0$ and $\boldsymbol{f}_i$ represent the drift and state-dependent control vectors, respectively.
Then, for the choice of $\boldsymbol{z} =  [{\phi}_1(\boldsymbol{x}),{\phi}_2(\boldsymbol{x}),...,{\phi}_{N}(\boldsymbol{x})]^{T}$, the corresponding Koopman representation is given as:
\begin{eqnarray}
    \label{eq: Koopman linear representation}
    \boldsymbol{\dot{z}} = \boldsymbol{A_c} \boldsymbol{z} + \boldsymbol{B}_{\boldsymbol{c}}  \boldsymbol{u}, \boldsymbol{x} = \boldsymbol{C}\boldsymbol{z}, 
\end{eqnarray}
where $\phi_i(\cdot):\mathbb{X} \to \mathbb{C} \; \forall  \; i =1,2, ..., N$ represents observable functions used to  arrive at the finite-dimensional approximation of the Koopman operator, with $\boldsymbol{A_c} \in \mathbb{R}^{N \times N}$ and
$\boldsymbol{B_c} \in \mathbb{R}^{N \times m}$  representing the corresponding Koopman matrices. $\boldsymbol{C} \in \mathbb{R}^{n \times N}$ represents a linear mapping from the lifted space state $\boldsymbol{z}$ to the base state of the system $\boldsymbol{x}$.
The corresponding discrete-time representations for the linear Koopman dynamics can be given as 
\begin{align}
    \label{eq:basic_lin}
    &\boldsymbol{{z}}_{k{+}1} {=} \boldsymbol{A}\boldsymbol{z}_k {+} \boldsymbol{B}\boldsymbol{u}_k, \,\, \boldsymbol{{x}}_{k} {=} \boldsymbol{C}\boldsymbol{{z}}_{k}.
\end{align}
A more in-depth analysis of the Koopman operator for control-affine systems can be found in \cite{goswami2017global, singh2024adaptive}. 

\subsection{Problem formulation}
Consider the noise-less dataset $\mathcal{D}_{a} = (\{\boldsymbol{x_{i}}\}_{i=1}^{N_s},\{\boldsymbol{x_{i}'}\}_{i=1}^{N_s},\{\boldsymbol{u_{i}}\}_{i=1}^{N_s}) $ , representing the system dynamics defined by (\ref{eq:base_affine}), where the $\{\boldsymbol{x_{i}'}\}_{i=1}^{N_s}$ denotes the one-step evolved states obtained by applying the corresponding control inputs $\{\boldsymbol{u_{i}}\}_{i=1}^{N_s}$ to the system at states $\{\boldsymbol{x_{i}}\}_{i=1}^{N_s}$. The associated noisy measurements acquired, are collected in $\mathcal{D}_m = (\{\boldsymbol{x_{mi}}\}_{i=1}^{N_s},\{\boldsymbol{x_{mi}'}\}_{i=1}^{N_s},\{\boldsymbol{u_{mi}}\}_{i=1}^{N_s})$. For each noisy state $\boldsymbol{x_{mi}}$, the lifted representation is defined as $\boldsymbol{z_{mi}} = \boldsymbol{\phi(x_{mi})}$,  where $\boldsymbol{\phi} = [\phi_1, \phi_2, \ldots, \phi_N]^{\top}$ denotes the dictionary of observable functions used for Koopman lifting.

Now, the Koopman operator $\boldsymbol{K_{fm}}$, which is approximated using available noisy data, is given as,
\begin{eqnarray}
\label{eq:koop_forward_noisy}
 \boldsymbol{K_{fm}} {=}\arg \min_{\boldsymbol{K_{m}}}\left\lVert{\begin{bmatrix}
                \boldsymbol{Z_m'}\\
                \boldsymbol{U_m}\end{bmatrix}} {-} \boldsymbol{K_m}{\begin{bmatrix}
                \boldsymbol{Z_m}\\
                \boldsymbol{U_m}
    \end{bmatrix}}\right\rVert_{F},
\end{eqnarray}
where $\boldsymbol{Z_m}$ is the snapshot matrix for the lifted states $\{\boldsymbol{z_{mi}} = \boldsymbol{\phi(x_{mi})}\}_{i=1}^{N_{s}}$ and $\boldsymbol{Z_m'}$ is the snapshot matrix for one-step ahead lifted state $\{\boldsymbol{z_{mi}'} = \boldsymbol{\phi(x_{mi}')}\}_{i=1}^{N_{s}}$ at the next timestep corresponding to $\boldsymbol{Z_m}$ when control input $\boldsymbol{u_m}$ acts over it. $\boldsymbol{U_m}$ is matrix with control inputs as its columns corresponding to states in $\boldsymbol{Z_m}$.

On the other hand, the Koopman operator computed using noise-free data is obtained as,
\begin{eqnarray}
 \boldsymbol{K_{f}} {=}\arg \min_{\boldsymbol{K}}\left\lVert{\begin{bmatrix}
                \boldsymbol{Z'}\\
                \boldsymbol{U}\end{bmatrix}} {-} \boldsymbol{K}{\begin{bmatrix}
                \boldsymbol{Z}\\
                \boldsymbol{U}
    \end{bmatrix}}\right\rVert_{F},
\end{eqnarray}
where $\boldsymbol{Z}$, $\boldsymbol{Z'}$ and $\boldsymbol{U}$ are noise-free matrices corresponding to $\boldsymbol{Z_m}$, $\boldsymbol{Z_m'}$ and $\boldsymbol{U_m}$.

Invoking linear least squares inversion as the solution of system (\ref{eq:koop_forward_noisy}) results in a biased estimate of the Koopman operator $\boldsymbol{K_{fm}}$ under noisy data, thus differing from the true noise-free Koopman operator $\boldsymbol{K_f}$. In practice, noise-free data is usually unavailable, leading to challenges in the accurate synthesis of the true Koopman operator from noisy data. Filtering noise may not always be suitable, as it may induce distortions, leading to data that represents altered dynamics rather than the true underlying system~\cite{sedehi-2025, Widmann2015DigitalFD}. To overcome these challenges, the goal of the current study is to propose a noise-robust Koopman architecture for the synthesis of the reduced-bias Koopman operator from noisy training data. 

 \begin{figure*}[h] 
\centering
    \includegraphics[width=\linewidth]{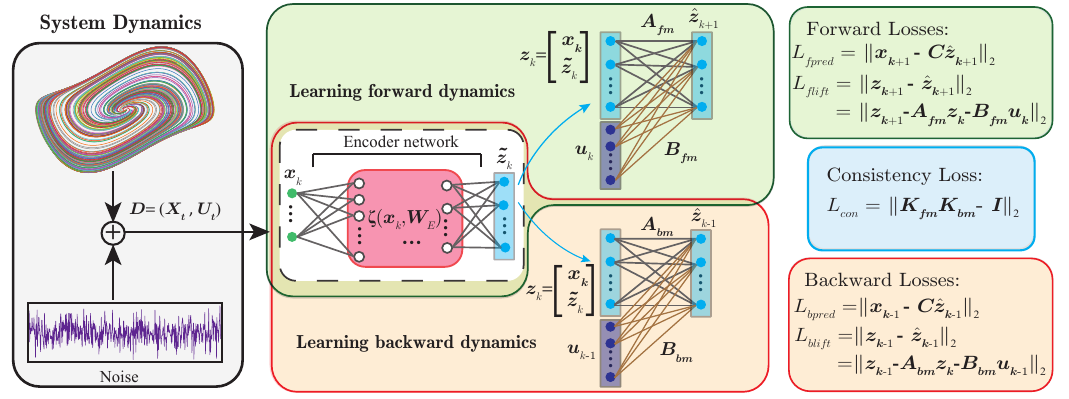}
    \caption{Schematics of the proposed DRKN architecture }
    \label{fig:DKRN_arch}
\end{figure*}

\section{Deep Robust Koopman Network}
In this section, we propose a noise-robust neural network architecture employed for the estimation of the finite-dimensional reduced-bias Koopman operator by leveraging the forward and backward temporal system dynamics.

 To begin, we define key terms related to the forward and backward dynamics as:
\begin{eqnarray}
    \label{eq:basic_fwd_bwd_lin}
    \boldsymbol{{z}}_{k{+}1}  {=}  \boldsymbol{A_{f}}\boldsymbol{z}_k {+} \boldsymbol{B_{f}}\boldsymbol{u}_k,
    \boldsymbol{{z}}_{k{-}1} {=} \boldsymbol{A_{b}}\boldsymbol{z}_k {+}
    \boldsymbol{B_{b}}\boldsymbol{u}_{k-1},
     \boldsymbol{{x}}_{k} {=} \boldsymbol{C}\boldsymbol{{z}}_{k},
\end{eqnarray}
where $\{\boldsymbol{A_f}$,$\boldsymbol{B_f}\}$ and $\{\boldsymbol{A_b}$,$\boldsymbol{B_b}\}$ represent the Koopman model matrices for the forward and backward temporal dynamics, respectively.

The architecture of the proposed Deep Robust Koopman Network (DRKN) is depicted in Fig. \ref{fig:DKRN_arch}. The proposed architecture, which incorporates an autoencoder to map the states to an observable space, is composed of two fully connected structurally identical linear layers that learn the forward and backward temporal dynamics of the system. This structure ensures that the autoencoder learns Koopman observables consistent across forward and backward temporal dynamics, reflecting the fact that the eigenfunctions spanning the Koopman subspace are invariant to the direction of flow of time. Consequently, the finite-dimensional representation of the Koopman operator for forward temporal dynamics is the inverse of that for backward temporal dynamics~\cite{9516947}. In this study, the original state $\boldsymbol{x_{k}}$ of the system is vertically stacked along with the nonlinear observables $\boldsymbol{\tilde{z}_k}$ learned by the auto-encoder to realize the lifted state $\boldsymbol{z_k} = [\boldsymbol{x_k},\boldsymbol{\tilde{z}_k}]^{\top}$, which is further propagated through linear layers to learn the forward and backward temporal dynamics. The decoder $\boldsymbol{C} {=} [\boldsymbol{I}_{n \times n}~~ \boldsymbol{0}] \in \mathbb{R}^{n \times N}$ is used to project the lifted states back to original states.
This architecture enables the neural network to simultaneously learn the basis functions and the Koopman operator by minimizing a loss function which comprises of the following terms: 



\begin{enumerate}
    \item \emph{Forward prediction loss} - This term signifies the difference in predicted and actual states for the forward temporal dynamics at the next time step, where predicted lifted states are estimated using the current weights of the fully connected forward linear layer and then projected to get back the predicted original states. It is given by  $ L_{fpred} = \boldsymbol{\|x_{k+1} {-} \hat{x}_{k+1}\|} = \boldsymbol{\| x_{k+1}  {-}  C(A_{fm}\zeta({x_k},W_E) {+} B_{fm}u_k)\| } $ where $ \boldsymbol{A_{fm}} \in \mathbb{R}^{N\times N}$ and $ \boldsymbol{B_{fm}} \in \mathbb{R}^{N\times m}$ are the matrix representations of weights of the fully connected forward linear layer as shown in Fig.~\ref{fig:DKRN_arch}.
    \item \emph{Forward lifted state prediction loss} - This term signifies the difference in predicted lifted observables and actual observables for the forward temporal dynamics at the next time step. It is given by $ L_{flift} = \boldsymbol{\|{z_{k+1}} {-} {\hat{z}_{k+1}}\|} = \boldsymbol{\|\zeta(x_{k+1},W_E) {-} A_{fm}\zeta(x_k,W_E) {-} B_{fm} u_k\| }$.
    \item \emph{Backward prediction loss} - This term quantifies the difference in actual and predicted states for the backward temporal dynamics at the previous time step. It is given by  $ L_{bpred} = \boldsymbol{\|x_{k-1} {-} \hat{x}_{k-1}\|} = \boldsymbol{\| x_{k-1}  {-}  C(A_{bm}\zeta({x_k},W_E) {+} B_{bm}u_{k-1})\| } $ where $ \boldsymbol{A_{bm}} \in \mathbb{R}^{N\times N}$ and $ \boldsymbol{B_{bm}} \in \mathbb{R}^{N\times m}$ are the matrix representations of weights of the fully connected backward linear layer as shown in Fig.~\ref{fig:DKRN_arch}.
    \item \emph{Backward lifted prediction loss} - This term represents the difference in actual and predicted lifted observables for the backward temporal dynamics at the previous time step. It is defined as $ L_{blift} =  \boldsymbol{\|{z_{k-1}} {-} {\hat{z}_{k-1}}\|} = \boldsymbol{\|\zeta(x_{k-1},W_E) {-} A_{bm}\zeta(x_k,W_E) {-} B_{bm} u_{k-1}\| }$ .
    \item \emph{Consistency loss} - 
    Consistency implies that the learned forward and backward Koopman operators are the inverses of each other for the same set of basis functions. It is quantified by the loss term $L_{con} = \boldsymbol{||K_{fm}K_{bm} - I ||}$, where $\boldsymbol{K_{fm}}$ and $\boldsymbol{K_{bm}}$ are defined by (\ref{eq:koop_for_noise}) and (\ref{eq:opt_prob_2}) respectively. The sole purpose of this loss term is to facilitate the learning of basis functions consistent to both forward and backward dynamics. 
\end{enumerate}

\begin{algorithm}[!t]
\caption{DRKN Koopman operator synthesis}
\label{alg:adaptive_koopman_mpc}
\begin{algorithmic}[1]

\REQUIRE  Dataset, $\boldsymbol{\mathcal{D}} = (\boldsymbol{X_t}, \boldsymbol{U_t})_{t=1}^T$, where T is number of trajectories.
\STATE Initialize the neural network parameters $\boldsymbol{\zeta(\cdot)}$, $\boldsymbol{A_{fm}}$, $\boldsymbol{B_{fm}}$, $\boldsymbol{A_{bm}}$, and $\boldsymbol{B_{bm}}$.

\FOR{each epoch $i$ in $1$ ... $E$}
\STATE Shuffle the batch order of the data
\FOR{each batch}
\STATE 1) Feed the batch data to get lifted state $\boldsymbol{z_{k}}$, $\boldsymbol{z_{k-1}}$ and $\boldsymbol{z_{k+1}}$.
\STATE 2) Compute the forward and backward predicted state $\boldsymbol{\hat{z}_{k+1}}$ and $\boldsymbol{\hat{z}_{k-1}}$ using (\ref{eq:basic_fwd_bwd_lin}).
\STATE 3) Compute the total loss $L(\boldsymbol{x_k},\boldsymbol{x_{k+1}},\boldsymbol{x_{k-1}},\boldsymbol{u_k})$
\STATE 4) Optimize for $\boldsymbol{\zeta(.)}$,$\boldsymbol{A_{fm}}$, $\boldsymbol{B_{fm}}$, $\boldsymbol{A_{bm}}$ and $\boldsymbol{B_{bm}}$.
\ENDFOR
\ENDFOR
\RETURN $\boldsymbol{A_{fm}, B_{fm}, A_{bm}, B_{bm}}$
\STATE Obtain the reduced-bias operator, $\boldsymbol{K_{proposed}}$, using (\ref{eq:ReducedBiasKoop}). 
\end{algorithmic}
\end{algorithm} 
Overall, the objective is to minimize the total loss, $ L(\boldsymbol{x_k,x_{k+1},x_{k-1},u_k}) = \alpha_1(L_{fpred} + L_{bpred}) {+} \alpha_2 (L_{flift} + L_{blift}) + \alpha_3 L_{con} +\gamma_1\|W_E\|_1 + \gamma_2\|W_E\|_2 $ over the whole dataset with $\alpha_1,\alpha_2,\alpha_3,\alpha_4,\gamma_1,\gamma_2 $ being the scalar weights.  This architecture utilizes this property of forward and backward flow to obtain the noise-robust Koopman operator from noisy data.

The forward Koopman operator, learned from noisy measured data, appears in block upper triangular form as:
\begin{eqnarray}
 \label{eq:koop_for_noise}
 \boldsymbol{K_{fm}} {=}\arg \min_{\boldsymbol{K_{f}}}\left\lVert{\begin{bmatrix}
                \boldsymbol{Z_m'}\\
                \boldsymbol{U_m}\end{bmatrix}} {-} \boldsymbol{K_f}{\begin{bmatrix}
                \boldsymbol{Z_m}\\
                \boldsymbol{U_m}
    \end{bmatrix}}\right\rVert_F {=} {\begin{bmatrix} \boldsymbol{A_{fm}}&\boldsymbol{B_{fm}} \\
 \boldsymbol{0}&\boldsymbol{I}\end{bmatrix}}.
\end{eqnarray}
Similarly, the backward Koopman operator is obtained as:
\begin{eqnarray}
    \label{eq:opt_prob_2}
    \boldsymbol{K_{bm}} = \arg \min_{\boldsymbol{K_b}}\left\lVert{\begin{bmatrix}
                \boldsymbol{Z_m}\\
                \boldsymbol{U_m}\end{bmatrix}} {-} \boldsymbol{K_b}{\begin{bmatrix}
                \boldsymbol{Z_m'}\\
                \boldsymbol{U_m}
    \end{bmatrix}}\right\rVert_{F} {=} {\begin{bmatrix} \boldsymbol{A_{bm}}&\boldsymbol{B_{bm}} \\
 \boldsymbol{0}&\boldsymbol{I}\end{bmatrix}}.
\end{eqnarray}
Then, the reduced bias forward Koopman operator can be written as:
\begin{align}
 \label{eq:ReducedBiasKoop}
 \boldsymbol{K_{proposed}} = \sqrt{\boldsymbol{K_{fm} K_{bm}^{-1}}} = \begin{bmatrix} \boldsymbol{A_{p}}&\boldsymbol{B_{p}} \\
 \boldsymbol{0}&\boldsymbol{I}\end{bmatrix}.
\end{align}
Observe that this reduced bias Koopman operator preserves the block upper triangular form consistent with forward and backward dynamics in (\ref{eq:koop_for_noise}) and (\ref{eq:opt_prob_2}). 

\section{Theoretical Analysis}
In this section, we undertake theoretical analysis regarding the noise robustness of the Koopman operator obtained using the proposed scheme. Let us consider finite sequences of randomly selected state vectors$\{\boldsymbol{x_{i}}\}_{i=1}^s$ and the finite sequence of control input as $\{\boldsymbol{u_i}\}_{i=1}^s$ acting on system (\ref{eq:base_affine}). Thus, we have associated sequence of states, $\{\boldsymbol{x_{i}^{+}}\}_{i=1}^s$ and $\{\boldsymbol{x_{i}^{-}}\}_{i=1}^s$, related as below,
\begin{align}
    \boldsymbol{x_i} = \boldsymbol{f_d(x_i^-,u_i)}, 
    \boldsymbol{x_i^+} = \boldsymbol{f_d(x_i,u_i)}, 
\end{align}
where $\boldsymbol{f_d}$ is discrete-time representation of (\ref{eq:base_affine}).

Let, the sequences denoted by $\{\boldsymbol{n_{xi}}\}_{i=1}^s$, $\{\boldsymbol{n_{xi}^{+}}\}_{i=1}^s$, $\{\boldsymbol{n_{xi}^{-}}\}_{i=1}^s$ and  $\{\boldsymbol{n_{ui}}\}_{i=1}^s$ define the noise components in the measurements of corresponding states of sequences $\{\boldsymbol{x_{i}}\}_{i=1}^s$, $\{\boldsymbol{x_{i}^{+}}\}_{i=1}^s$, $\{\boldsymbol{x_{i}^{-}}\}_{i=1}^s$ and $\{\boldsymbol{u_i}\}_{i=1}^s$ respectively, such that $\{\boldsymbol{x_{mi}}\}_{i=1}^s = \{\boldsymbol{x_i}\}_{i=1}^s + \{\boldsymbol{n_{xi}}\}_{i=1}^s,\,\{\boldsymbol{x_{mi}^{+}}\}_{i=1}^s = \{\boldsymbol{x_i^{+}}\}_{i=1}^s + \{\boldsymbol{n_{xi}^{+}}\}_{i=1}^s,\,\{\boldsymbol{x_{mi}^{-}}\}_{i=1}^s = \{\boldsymbol{x_i^{-}}\}_{i=1}^s + \{\boldsymbol{n_{xi}^{-}}\}_{i=1}^s,\,\{\boldsymbol{u_{mi}}\}_{i=1}^s = \{\boldsymbol{u_i}\}_{i=1}^s + \{\boldsymbol{n_{ui}}\}_{i=1}^s$. 
Note that the noise is assumed to be normally distributed with zero mean (i.i.d). Now the elements of each sequence can be stacked horizontally to generate the corresponding matrices defined below.
\begin{align}
    &\boldsymbol{X}    = [\boldsymbol{x_1}, \; \boldsymbol{x_2}, \dots,  \; \boldsymbol{x_{s}} ],\,\ \boldsymbol{X^{+}} = [\boldsymbol{x_1^{+}}, \; \boldsymbol{x_2^{+}}, \dots,  \; \boldsymbol{x_{s}^{+}} ],\,\nonumber \\
    &\boldsymbol{X^{-}} = [\boldsymbol{x_1^{-}}, \; \boldsymbol{x_2^{-}}, \dots,  \; \boldsymbol{x_{s}^{-}} ],\,\ \boldsymbol{U} = [\boldsymbol{u_1}, \; \boldsymbol{u_2}, \dots,  \; \boldsymbol{u_{s}} ],\,\ \nonumber
\end{align}
where $\boldsymbol{X}$, $\boldsymbol{X^{+}}$, $\boldsymbol{X^{-}}$ $ \in \mathbb{R}^{n \times s}$ and $\boldsymbol{U}$ $ \in \mathbb{R}^{m \times s}$ denote matrices corresponding to the true system states and matrix of inputs respectively. The matrices $\boldsymbol{N_x}=\boldsymbol{(n_{xi})_{i=1}^s}$, $\boldsymbol{N_x^{+} =(n_{xi}^{+})_{i=1}^s}$, $\boldsymbol{N_x^{-} = (n_{xi}^{-})_{i=1}^s }$ $ \in \mathbb{R}^{n \times s}$ represent the noise components associated with these state matrices. Consequently, the measured (noisy) state snapshot matrices are given by $\boldsymbol{X_m} = \boldsymbol{(x_{mi})_{i=1}^{s}} = \boldsymbol{X} + \boldsymbol{N_x}$, $\boldsymbol{X_m^{+}} = \boldsymbol{(x_{mi}^{+})_{i=1}^{s}} = \boldsymbol{X^{+}} + \boldsymbol{N_x^{+}}$, $\boldsymbol{X_m^{-}} = \boldsymbol{(x_{mi}^{-})_{i=1}^{s}} = \boldsymbol{X^{-}} + \boldsymbol{N_x^{-}}$ and measured inputs snapshot matrix is represented by $\boldsymbol{U_{m}} = \boldsymbol{U} + \boldsymbol{{N_{u}}}$. The lifted state matrix of the measured states is denoted as $\boldsymbol{Z_m} = \boldsymbol{Z} + \boldsymbol{N_z}$, where $\boldsymbol{Z}$ $ \in \mathbb{R}^{N \times s}$  corresponds to the lifted representation of the true states, and $\boldsymbol{N_z} = [\boldsymbol{\phi(x_{m1})-\phi(x_1)},    \boldsymbol{\phi({x_{m2}})-\phi(x_2)},\dots,\;\boldsymbol{\phi(x_{ms})-\phi(x_{s})}]$ $ \in \mathbb{R}^{N \times s}$  captures the associated noise in the lifted space. Similarly, the lifted measurement snapshot matrices corresponding to $\boldsymbol{Z^{+}}$ and $\boldsymbol{Z^{-}}$ are denoted by $\boldsymbol{Z_{m}^{+}}$ and $\boldsymbol{Z_{m}^{-}}$, respectively, with $\boldsymbol{N_{z}^{+}}$ and $\boldsymbol{N_{z}^{-}}$ representing the associated noise in the lifted space.

Now, the biased forward Koopman operator, $\boldsymbol{K_{fm}}$ is obtained by directly solving the following optimization problem using noisy measured data.
\begin{eqnarray}
    \label{eq:opt_prob}
    \boldsymbol{K_{fm}}&=&\arg \min_{\boldsymbol{K_m}}||\boldsymbol{\Psi_{m}^{+}} {-} \boldsymbol{K_m}\boldsymbol{\Psi_{m}}||_{F}\nonumber\\
    \boldsymbol{\Psi_{m}^{+}}&=&\begin{bmatrix}
                \boldsymbol{Z_m^{+}}\\
                \boldsymbol{U_m}
    \end{bmatrix}, \;
    \boldsymbol{\Psi_{m}} {=} \begin{bmatrix}
                \boldsymbol{Z_m}\\
                \boldsymbol{U_m}
    \end{bmatrix}, \; \boldsymbol{\Psi_{m}^{-}} {=} \begin{bmatrix}
                \boldsymbol{Z_m^{-}}\\
                \boldsymbol{U_m}
    \end{bmatrix}.
\end{eqnarray}
Similary, the backward Koopman operator $\boldsymbol{K_{bm}}$ can be obtained by solving the following optimization problem:
\begin{align}
\label{eq:opt_prob_backward}
    \boldsymbol{K_{bm}} = &\arg \min_{\boldsymbol{K_m}}||\boldsymbol{\Psi_{m}^{-}} - \boldsymbol{K_m}\boldsymbol{\Psi_{m}}||_F,
\end{align}
where $\boldsymbol{\Psi_{m}^{-}}$, $\boldsymbol{\Psi_m}$, $\boldsymbol{\Psi_m^{+}}\in \mathbb{R}^{(N+m)\times s}$ can be further expanded as $\boldsymbol{\Psi_m^{-}} = \boldsymbol{\Psi^{-}} + \boldsymbol{N_{\psi}^{-}}$, $\boldsymbol{\Psi_m} = \boldsymbol{\Psi} + \boldsymbol{N_{\psi}}$ and $\boldsymbol{\Psi_m^{+}} = \boldsymbol{\Psi^{+}} + \boldsymbol{N_{\psi}^{+}}$ respectively. The following assumptions are now considered for further analysis:

{
\itshape
Assumption 1-\textnormal{\cite{singh2024adaptive}}: The finite set of basis functions $ \{\phi_1, ..., \phi_N\}$ learned through the autoencoder, along with the inputs $ \{u_1, ..., u_{s}\}$ are assumed to span the Koopman-invariant subspace.

Assumption 2-\textnormal{\cite{Dawson2014CharacterizingAC}}: The system dynamics are assumed to be temporally invertible i.e $\boldsymbol{K_b = K_{f}}^{-1}$ exists, where $\boldsymbol{K_{f}} = \boldsymbol{\Psi^{+}}\boldsymbol{\Psi}^\top(\boldsymbol{\Psi}\boldsymbol{\Psi}^\top)^{-1}$ is the true Koopman operator computed for the underlying noiseless system. 

Assumption 3-\textnormal{\cite{annurev:/content/journals/10.1146/annurev-control-071020-010108}
} The norms $\boldsymbol{||K_f||\approx ||K_b|| \approx 1}$.

Assumption 4 - The noise levels are sufficiently small such that the following inequality holds,
\begin{eqnarray}
\label{eq:noise_level_assump}
\boldsymbol{|| (\Psi_i\Psi_j^{\top})^{-1}||\,||N_{\psi_i}\Psi_i^\top {+} \Psi_k N_{\psi_j}^\top {+} N_{\psi_i}N_{\psi_j}^\top || \ll 1},
\end{eqnarray} 
$\forall \boldsymbol{\Psi_i,\Psi_j,\Psi_k}\in{\boldsymbol{\Psi^{-},\Psi,\Psi^{+}}},\, \boldsymbol{N_{\psi_i}, N_{\psi_j}} \in{\boldsymbol{N_{\psi}^{-}, N_{\psi},}\boldsymbol{N_{\psi}^{+}}}$.
}

\textit{Remark 1:}
Assumption 1 imposes a mild condition on the synthesis of the linear Koopman model, similar to that considered in \cite{singh2024adaptive} and \cite{sah2024real}. Assumption 2 considers the system dynamics to be temporally invertible, which is true for most physical systems~\cite{Dawson2014CharacterizingAC}. Assumption 3 concerns the norm of the Koopman operator, which is known to hold for measure-preserving conservative dynamical systems, as demonstrated in \cite{annurev:/content/journals/10.1146/annurev-control-071020-010108}. In systems with control inputs, the operator norm may slightly exceed unity. However, the primary purpose of this assumption is to establish an approximate order of magnitude for the Koopman operator norm, which facilitates the subsequent analysis. Assumption 4 concerns the relative magnitudes of the noise levels and the system states in the lifted space. Owing to the structure of the autoencoder employing bounded activation functions, together with finite weights learned by the output layer of the autoencoder network, it follows that the lifting function $\boldsymbol{\phi}$ is globally Lipschitz and bounded\cite{szegedy2014intriguingpropertiesneuralnetworks}. Under these conditions, the norm of matrices $\boldsymbol{N_{\psi}, N_{\psi}^{+}}$ and $\boldsymbol{N_{\psi}^{-}}$ which contain lifted and one time-step evolved lifted noise components are approximately of the same order, i.e $\boldsymbol{||N_\psi|| \approx ||N_\psi^{+}|| \approx ||N_\psi^{-}||}$. This ensures that the inequalities specified in Assumption 4 are satisfied for all admissible combinations of noise and state matrices.

We now investigate the impact of noise on the learning of the Koopman operator. The proposed autoencoder-based neural architecture facilitates the automatic selection of Koopman basis functions through the minimization of an appropriate loss function. Consequently, once the autoencoder converges and the lifting functions are learned, the subsequent linear layer identifies the Koopman operator by effectively solving the optimization problem in (\ref{eq:opt_prob}). 

For  $\boldsymbol{\Psi_{m}}\in \mathbb{R}^{(N+m)\times s}$ with $ s> (N+m)$, the solution to (\ref{eq:opt_prob}) can be obtained by invoking the Moore-Penrose pseudoinverse $\boldsymbol{{\Psi_{m}}^{\dagger}}{=}\boldsymbol{\Psi_{m}^{\top}}(\boldsymbol{\Psi_{m}}\boldsymbol{\Psi_{m}}^{\top})^{-1}$ as:
\begin{align}
 \label{eq:noisy_koop_exp}
    \boldsymbol{K_{fm}} =& \boldsymbol{{\Psi_{m}^{+}} {\Psi_{m}}^{\dagger}}{=}({\boldsymbol{\Psi^{+}} + \boldsymbol{N_\psi^{+}}} )({\boldsymbol{\Psi} + \boldsymbol{N_\psi}})^{\dagger}, \nonumber \\
    = &({\boldsymbol{\Psi^{+}} + \boldsymbol{N_\psi^{+}}})({\boldsymbol{\Psi} + \boldsymbol{N_\psi}})^{\top} \bigl(({\boldsymbol{\Psi} + \boldsymbol{N_\psi}})
    ({\boldsymbol{\Psi} + \boldsymbol{N_\psi}})^{\top}\bigr)^{-1},\nonumber\\
    = &(\boldsymbol{\Psi^{+}}\boldsymbol{\Psi}^\top + \boldsymbol{N_\psi^{+}}\boldsymbol{\Psi}^{\top} + \boldsymbol{\Psi^{+}}\boldsymbol{N_\psi}^\top + \boldsymbol{N_\psi^{+}}\boldsymbol{N_\psi}^\top) 
     (\boldsymbol{\Psi}\boldsymbol{\Psi}^\top + \nonumber \\
    &\boldsymbol{N_\psi}\boldsymbol{\Psi}^{\top} + \boldsymbol{\Psi}\boldsymbol{N_\psi}^\top + \boldsymbol{N_\psi}\boldsymbol{N_\psi}^\top)^{-1}.
\end{align}
For every $p \in \mathbb{N}$ and $\boldsymbol{M},\,\boldsymbol{P} \in \mathbb{R}^{p \times p}$ satisfying $||\boldsymbol{M^{-1}P}||\ll1$, it follows from \cite{Dawson2014CharacterizingAC},\cite{golub13} that
\begin{align}
\label{eq:Matrix_perturb_inv_expansion}
(\boldsymbol{M} + \boldsymbol{P})^{-1} \approx  ~&\boldsymbol{M}^{-1} - \boldsymbol{M}^{-1}\boldsymbol{P}\boldsymbol{M}^{-1}.\end{align}
Then, invoking (\ref{eq:noise_level_assump}) with the fact that $||\boldsymbol{M}^{-1}\boldsymbol{P}||< ||\boldsymbol{M}^{-1}||\,||\boldsymbol{P}||$ and using (\ref{eq:Matrix_perturb_inv_expansion}), we obtain,
\begin{eqnarray}
  \label{eq:forward_koop_approx_1}
  \boldsymbol{K_{fm}} {=}(\boldsymbol{\Psi^{+}}\boldsymbol{\Psi}^{\top} {+} \boldsymbol{N_\psi^{+}}\boldsymbol{\Psi}^{\top} {+} \boldsymbol{\Psi^{+}}\boldsymbol{N_\psi}^\top {+} \boldsymbol{N_\psi^{+}}\boldsymbol{N_\psi}^{\top})\qquad\nonumber\\\times(\boldsymbol{\Psi}\boldsymbol{\Psi}^\top)^{-1}(\boldsymbol{I}{-} \boldsymbol{P_1}) {=}\underbrace{\boldsymbol{K_f}(\boldsymbol{I}{-} \boldsymbol{P_1})}_{T_1}\nonumber\\{+}\underbrace{(\boldsymbol{N_\psi^{+}}\boldsymbol{\Psi}^{\top}+\boldsymbol{\Psi^{+}}\boldsymbol{N_\psi}^\top + \boldsymbol{N_\psi^{+}}\boldsymbol{N_\psi}^\top) (\boldsymbol{\Psi}\boldsymbol{\Psi}^{\top})^{-1}}_{\boldsymbol{T_2}}\,\,\,\,\,\nonumber\\
    {-}\underbrace{(\boldsymbol{N_\psi^{+}}\boldsymbol{\Psi} +\boldsymbol{\Psi^{+}}\boldsymbol{N_\psi}^\top +\boldsymbol{N_\psi^{+}}\boldsymbol{N_\psi}^\top)(\boldsymbol{\Psi}\boldsymbol{\Psi}^{\top})^{-1}\boldsymbol{P_1}}_{\boldsymbol{T_3}}.
\end{eqnarray}
where $\boldsymbol{P_1} = (\boldsymbol{N_{\psi}\Psi^\top + \Psi N_{\psi}^\top + N_{\psi}N_{\psi}^\top})(\boldsymbol{\Psi}\boldsymbol{\Psi^\top})^{-1}$ and $\boldsymbol{K_f} = \boldsymbol{\Psi^{+}}\boldsymbol{\Psi}^\top(\boldsymbol{\Psi}\boldsymbol{\Psi}^\top)^{-1} $ is the forward Koopman operator for the underlying noise-less system. Now, using matrix perturbation theory, under Assumption 4, $\boldsymbol{T_3}$ is the product of two small terms and is therefore negligible, yielding $||\boldsymbol{T_3}||\ll ||\boldsymbol{T_1}||$. Hence, ignoring $\boldsymbol{T_3}$,
\begin{align}
\label{eq:Koopman_fwd_noise}
    \boldsymbol{K_{fm}} \approx\boldsymbol{K_f(I-P_1)} + \boldsymbol{T_2}
    =\boldsymbol{K_f}\boldsymbol{\mathcal{B}} + \boldsymbol{\Delta_f},
\end{align}
where $\boldsymbol{\mathcal{B}} = \boldsymbol{I} - \boldsymbol{P_1}$ and $\boldsymbol{\Delta_f} = \boldsymbol{T_2}$.
Since, $\boldsymbol{K_{fm}}$ consists of statistical elements related to noise, we compute the expectation value as $\mathbb{E}[\boldsymbol{K_{fm}}] \approx\boldsymbol{K_f}\mathbb{E}[\boldsymbol{\mathcal{B}}] + \mathbb{E}[\boldsymbol{\Delta_f}]$,
where $\mathbb{E}[\boldsymbol{\mathcal{B}}] = \boldsymbol{I} - \mathbb{E}[\boldsymbol{P_1}] = \boldsymbol{I} - (\mathbb{E}[\boldsymbol{N_{\psi}}]\boldsymbol{\Psi^\top} + \boldsymbol{\Psi}\mathbb{E}[\boldsymbol{N_{\psi}^{\top}}]+\mathbb{E}[\boldsymbol{N_\psi N_\psi^\top}])(\boldsymbol{\Psi\Psi^{\top}})^{-1}$ represents the multiplicative bias term affecting $\boldsymbol{K_f}$ induced by noise. Importantly, unlike the linear basis functions that DMD relies on for lifting \cite{Dawson2014CharacterizingAC}, note that $\mathbb{E}[\boldsymbol{N_\psi}]\neq 0$ since noise is lifted through nonlinear basis functions in this study. A similar conclusion as (\ref{eq:Koopman_fwd_noise}) can be obtained for the case of learning the backward Koopman operator \eqref{eq:opt_prob_backward} as,
\begin{align}
    \boldsymbol{K_{bm}}\approx\boldsymbol{K_b}\boldsymbol{\mathcal{B}} + \boldsymbol{\Delta_b}\label{eq:Koopman_bwd_noise},
\end{align}
and $\mathbb{E}[\boldsymbol{K_{bm}}]\approx\boldsymbol{K_b}\mathbb{E}[\boldsymbol{\mathcal{B}}] + \mathbb{E}[\boldsymbol{\Delta_b}]$, where $\boldsymbol{\Delta_b} = (\boldsymbol{N_{\psi}^{-}\Psi^\top + \Psi^{-} N_{\psi}^\top + N_{\psi}^{-}N_{\psi}^\top})(\boldsymbol{\Psi}\boldsymbol{\Psi^\top})^{-1}$. Then, much like \cite{Dawson2014CharacterizingAC}, 
as the noise statistics in the lifted space remains unknown, it is more straightforward to work with the expressions for $\boldsymbol{K_{fm}}$ and $\boldsymbol{K_{bm}}$ in (\ref{eq:Koopman_fwd_noise}) and (\ref{eq:Koopman_bwd_noise}) instead. In particular, in order to cancel out the effects of the bias term $\boldsymbol{\mathcal{B}}$ affecting the nominal Koopman operator, the noise-robust (reduced-bias) Koopman operator is proposed as 
\begin{eqnarray}
    \label{eq:proposed_koop}
    \boldsymbol{K_{proposed}} = \sqrt{\boldsymbol{K_{fm}}\boldsymbol{K_{bm}}^{-1}}.
\end{eqnarray}

Estimating the forward Koopman operator using the inverse of the backward operator as $\boldsymbol{K_{fm}^{back}} = \boldsymbol{K_{bm}^{-1}}$ leads to the inversion of the bias term in (\ref{eq:Koopman_bwd_noise}). So, the multiplication of the forward and backward dynamics in (\ref{eq:proposed_koop}) tends to cancel the significant part of noise-induced bias, resulting in a reduced bias estimate of the Koopman operator. However, it should be noted that the presence of the terms $\boldsymbol{\Delta_f}$ and $\boldsymbol{\Delta_b}$, in \eqref{eq:Koopman_fwd_noise} and \eqref{eq:Koopman_bwd_noise} respectively, leads to a persistent residual bias in the proposed operator $\boldsymbol{K_{proposed}}$ defined in \eqref{eq:proposed_koop}. 

It then becomes important to show that the proposed noise-robust Koopman operator $\boldsymbol{K_{proposed}} =\sqrt{\boldsymbol{K_{{fm}} K_{{bm}}^{-1}}} $ satisfies the inequality:
\begin{eqnarray}    
    \label{robust_Koopman_inequality}
    ||\mathbb{E}[\boldsymbol{K_{proposed}^2}]-\boldsymbol{K_f^{2}}|| < ||\mathbb{E}[\boldsymbol{K_{fm}^2}]-\boldsymbol{K_f^{2}}||.
\end{eqnarray}

To this end, invoking Assumption 4, and using \eqref{eq:Matrix_perturb_inv_expansion}, we obtain 
\begin{align}
    \boldsymbol{K_{bm}}^{-1} = \boldsymbol{\mathcal{B}}^{-1}\boldsymbol{K_b}^{-1} - \boldsymbol{\mathcal{B}}^{-1}\boldsymbol{K_b}^{-1}\boldsymbol{\Delta_b}\boldsymbol{\mathcal{B}}^{-1}\boldsymbol{K_b}^{-1}.\label{eq:Koopman_fwd_using_bwd}
\end{align}
Multiplying \eqref{eq:Koopman_fwd_noise} and \eqref{eq:Koopman_fwd_using_bwd} we have,
\begin{align}
\label{eq:Koopman_fwd_bwd_expanded}
    \boldsymbol{K_{fm}K_{bm}}^{-1} = &~\boldsymbol{K_f}^{2} - \boldsymbol{K_f}^2\boldsymbol{\Delta_b}\boldsymbol{\mathcal{B}}^{-1}\boldsymbol{K_b}^{-1} + \boldsymbol{\Delta_f}\boldsymbol{\mathcal{B}}^{-1}\boldsymbol{K_b}^{-1} \nonumber\\&- \boldsymbol{\Delta_f}\boldsymbol{\mathcal{B}}^{-1}\boldsymbol{K_b}^{-1}\boldsymbol{\Delta_b}\boldsymbol{\mathcal{B}}^{-1}\boldsymbol{K_b}^{-1}.
\end{align}
Invoking Assumption 4 and using \eqref{eq:Matrix_perturb_inv_expansion} we have $\boldsymbol{\mathcal{B}}^{-1} = (\boldsymbol{I} - \boldsymbol{P_1})^{-1} \approx (\boldsymbol{I+P_1})$. Thus, we obtain from \eqref{eq:Koopman_fwd_bwd_expanded},
\begin{align}
    \boldsymbol{K_{fm}K_{bm}}^{-1}=&~\boldsymbol{K_f}^{2} - \boldsymbol{K_f}^2\boldsymbol{\Delta_b}\boldsymbol{K_b}^{-1} - \boldsymbol{K_f}^2\boldsymbol{\Delta_b P_1}\boldsymbol{K_b}^{-1}\nonumber \\&+ \boldsymbol{\Delta_f}\boldsymbol{K_b}^{-1} + \boldsymbol{\Delta_f P_1}\boldsymbol{K_b}^{-1} - \boldsymbol{\Delta_f}\boldsymbol{K_b}^{-1}\boldsymbol{\Delta_b}\boldsymbol{K_b}^{-1}\nonumber\\& - \boldsymbol{\Delta_f P_1}\boldsymbol{K_b}^{-1}\boldsymbol{\Delta_b}\boldsymbol{K_b}^{-1} - \boldsymbol{\Delta_f}\boldsymbol{K_b}^{-1}\boldsymbol{\Delta_b P_1}\boldsymbol{K_b}^{-1}\nonumber\\& - \boldsymbol{\Delta_f P_1}\boldsymbol{K_b}^{-1}\boldsymbol{\Delta_b P_1}\boldsymbol{K_b}^{-1}.\label{eq:Koopman_fwd_bwd_full_exp}
\end{align}
Under Assumption 4, observe that $||\boldsymbol{P_1}||\approx||\boldsymbol{\Delta_f}||\approx||\boldsymbol{\Delta_b}||\ll1$. Thus, the higher order terms in \eqref{eq:Koopman_fwd_bwd_full_exp}, consisting of a product of two or more such matrices, become negligible, so that, using $\boldsymbol{K_f}=\boldsymbol{K_b}^{-1}$,
\begin{eqnarray}
\label{eq:Koopman_fwd_bwd_final}
    \boldsymbol{K_{fm}K_{bm}}^{-1}-\boldsymbol{K_f}^{2}{\approx} {-} \boldsymbol{K_f}^2\boldsymbol{\Delta_b}\boldsymbol{K_f} {+} \boldsymbol{\Delta_f}\boldsymbol{K_f}{=} \boldsymbol{S_1}\boldsymbol{K_f},\label{eq:Koopman_fwd_bwd_exp_short}
\end{eqnarray}
where $\boldsymbol{S_1} = (\boldsymbol{N_{\psi}^{+}}{-}\boldsymbol{K_f^{2}}\boldsymbol{N_{\psi}^{-}})(\boldsymbol{\Psi^{\top}}{+}\boldsymbol{N_{\psi}^{\top}})(\boldsymbol{\Psi}\boldsymbol{\Psi^{\top}})^{-1} $.
Now, using $\boldsymbol{\mathcal{B}} = \boldsymbol{I-P_1}$ and \eqref{eq:Koopman_fwd_noise},
\begin{align}
    \boldsymbol{K_{fm}}^{2}-\boldsymbol{K_{f}}^{2} \approx~(\boldsymbol{K_f}\boldsymbol{\mathcal{B}}+\boldsymbol{\Delta_f})(\boldsymbol{K_f}\boldsymbol{\mathcal{B}}+\boldsymbol{\Delta_f})-\boldsymbol{K_{f}}^{2}\nonumber\\
    ={-} \boldsymbol{K_f P_1}\boldsymbol{K_f} {-} \boldsymbol{K_f}^{2}\boldsymbol{P_1} {+} \boldsymbol{K_f P_1}\boldsymbol{K_f P_1} {+} \boldsymbol{K_f}\boldsymbol{\Delta_f}\nonumber\\ {-} \boldsymbol{K_f P_1}\boldsymbol{\Delta_f} {+} \boldsymbol{\Delta_f}\boldsymbol{K_f} {-} \boldsymbol{\Delta_f}\boldsymbol{K_f P_1} {+} \boldsymbol{\Delta_f}^{2}.\label{eq:Koopman_fwd_sq_exp}
\end{align}
Similarly, the higher order terms in \eqref{eq:Koopman_fwd_sq_exp}, involving the product of two or more such matrices, become negligible, so that,
\begin{eqnarray}
        \boldsymbol{K_{fm}^2} {-} \boldsymbol{K_f}^2{\approx}(\boldsymbol{\Delta_f}{-} \boldsymbol{K_f}\boldsymbol{P_1})\boldsymbol{K_f} {+} \boldsymbol{K_f}(\boldsymbol{\Delta_f} {-} \boldsymbol{K_f P_1})\qquad\nonumber\\
{=}[(\boldsymbol{N_{\psi}^{+}}-\boldsymbol{K_f}\boldsymbol{N_{\psi}})\boldsymbol{\Psi^{\top}}{+}(\boldsymbol{\Psi^{+}}-\boldsymbol{K_f}\boldsymbol{\Psi})\boldsymbol{N_{\psi}^{\top}}\qquad\qquad\nonumber\\{+}(\boldsymbol{N_{\psi}^{+}}-\boldsymbol{K_f}\boldsymbol{N_{\psi}})(\boldsymbol{N_{\psi}^{\top}})](\boldsymbol{\Psi}\boldsymbol{\Psi^{\top}})^{-1}\boldsymbol{K_f}\qquad\qquad\qquad\nonumber \\
        {+}\boldsymbol{K_f}[(\boldsymbol{N_{\psi}^{+}}{-}\boldsymbol{K_f}\boldsymbol{N_{\psi}})\boldsymbol{\Psi^{\top}}{+}(\boldsymbol{\Psi^{+}}{-}\boldsymbol{K_f}\boldsymbol{\Psi})\boldsymbol{N_{\psi}^{\top}}\qquad\qquad\nonumber\\
        {+}(\boldsymbol{N_{\psi}^{+}}-\boldsymbol{K_f}\boldsymbol{N_{\psi}})(\boldsymbol{N_{\psi}^{\top}})](\boldsymbol{\Psi}\boldsymbol{\Psi^{\top}})^{-1}{=}\boldsymbol{S_2}\boldsymbol{K_f} {+} \boldsymbol{K_f}\boldsymbol{S_2},
        \label{eq:Koopman_fwd_expanded}
\end{eqnarray}
where, we have used the fact that $\boldsymbol{\Psi^{+}}=\boldsymbol{K_f}\boldsymbol{\Psi}$, and $\boldsymbol{S_2} = (\boldsymbol{N_{\psi}^{+}}{-}\boldsymbol{K_f}\boldsymbol{N_{\psi}})(\boldsymbol{\Psi^{\top}}{+}\boldsymbol{N_{\psi}^{\top}})(\boldsymbol{\Psi}\boldsymbol{\Psi^{\top}})^{-1}$. In \eqref{eq:Koopman_fwd_bwd_exp_short} and \eqref{eq:Koopman_fwd_expanded}, the terms $\boldsymbol{S_1}$ and $\boldsymbol{S_2}$ include random noise components that are lifted via nonlinear functions, and therefore their statistical properties cannot be explicitly characterized. However, we leverage the fact that the discrete-time Koopman operator, $\boldsymbol{K_f}$, can be written as,
\begin{align}
    \boldsymbol{K_f} =& \boldsymbol{e^{\mathcal{K}\Delta t}}=\boldsymbol{I}+\underbrace{\boldsymbol{\mathcal{K}}\Delta t}_{\boldsymbol{R}} + \mathcal{O}(\Delta t^2),\label{eq:cont-discrete-time_Koopman}
\end{align}
where $\Delta t\,(\ll1)$ is the sampling time, and $\boldsymbol{\mathcal{K}}$ is the (unknown) ideal continuous-time Koopman representation corresponding to the noiseless system (\ref{eq: Koopman linear representation}). Hence, for a sufficiently small sampling interval, we ignore the higher order terms in \eqref{eq:cont-discrete-time_Koopman}, and have $\boldsymbol{K_f} \approx \boldsymbol{I} + \boldsymbol{R}$ with $\boldsymbol{||R||}\ll1$. As $\boldsymbol{||N_\psi|| \approx ||N_\psi^{+}|| \approx ||N_\psi^{-}||}$, it is then straightforward to verify that $||\mathbb{E}[\boldsymbol{S_1}]||\approx ||\mathbb{E}[\boldsymbol{S_2}]||\approx\delta$ for a constant $\delta>0$, so that,
\begin{eqnarray}
    \mathbb{E}[\boldsymbol{K_{fm}^{2}] {-} K_f^{2}} {\approx}~2\mathbb{E}[\boldsymbol{S_2}],\,
    \mathbb{E}[\boldsymbol{K_{fm}K_{bm}}^{-1}]{-}\boldsymbol{K_f}^{2} {\approx} ~\mathbb{E}[\boldsymbol{S_1}].
\end{eqnarray}
Thus, we obtain,
\begin{eqnarray}
\label{eq:Koopman_proposed}
    ||\mathbb{E}[\boldsymbol{K_{proposed}^2}] {-} \boldsymbol{K_f}^2|| \approx{~}\delta,\,
    ~||\mathbb{E}[\boldsymbol{K_{fm}^2}] {-} \boldsymbol{K_f}^2 || {\approx} ~ 2\delta.
\end{eqnarray}
The inequality in (\ref{robust_Koopman_inequality}) then follows directly, so that employing the proposed Koopman operator reduces the bias induced by noise. Clearly, incorporating both forward and backward temporal dynamics mitigates noise-induced bias in the learned Koopman operator without requiring explicit estimation of the noise statistics corrupting the lifted state, which is an important feature of the proposed scheme.

\textit{Remark 2:}
The theoretical analysis presented above extends the forward-backward DMD framework introduced in \cite{Dawson2014CharacterizingAC}, which was restricted to autonomous systems with original states as Koopman observables. In contrast, the present analysis accommodates nonlinear Koopman observables for non-autonomous systems in control-affine form, so that it reduces to the formulation in \cite{Dawson2014CharacterizingAC} for the special case when the system is autonomous and the states are directly considered as Koopman observables. It is crucial to note that the synthesis of the Koopman operator with a reduced bias relies on the satisfaction of inequality (\ref{eq:noise_level_assump}), which may not strictly hold for high noise levels. However, the simulation results in the following section clearly demonstrate that the proposed architecture remains robust and effectively mitigates the noise-induced bias even for relatively high levels of measurement noise, which is an important contribution of the proposed scheme. 
 \begin{figure*}[h] 
\centering
    \includegraphics[width=\linewidth]{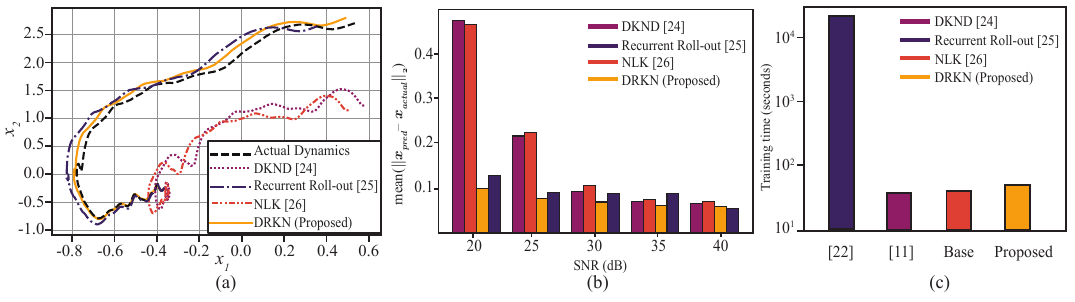}
    \caption{\footnotesize Performance comparison for learning the dynamics of a Van der Pol oscillator a) Phase portrait generated using learned Koopman models from data with SNR 20 dB. b) Predictive performance comparison. c) Training time comparison with a dataset corrupted by 20dB. noise.}
    \label{fig:van_pol_sim}
\end{figure*}
 \section{Simulation Results}
This section presents comprehensive simulation studies to evaluate the effectiveness of the proposed DRKN architecture in learning linear Koopman models from noisy data. We consider two representative nonlinear systems: the Van der Pol oscillator and serial manipulators. The DRKN is first trained on datasets corrupted with measurement noise to emulate real-world conditions. To assess the quality of the learned model, its prediction performance is evaluated on noise-free trajectories, isolating model accuracy from noise effects. For closed-loop trajectory tracking, however, feedback is assumed to be noisy to reflect realistic sensor conditions. Additionally, we compare the performance of DRKN against leading state-of-the-art approaches, including DKND~\cite{Hao2024DeepKL}, Recurrent Roll-out~\cite{Sakib2024LearningNS}, and the nominal Koopman (NLK)~\cite{singh2024adaptive}, demonstrating the superior accuracy and robustness of the proposed method. Finally, the efficacy of the learnt model in achieving precise closed-loop trajectory tracking is also demonstrated across various manipulator systems. 
\subsection{Van der Pol oscillator}
 Consider the Van der Pol oscillator with control input:
 \begin{align}    
    \dot{x}_1 {=}-{x_2},\,
    \dot{x}_2 {=} \mu(-1 + {x_1}^2)x_2 + x_1 + u
\end{align}
where $u$ is the control input and $\mu = 1$. For collecting data, the time step, $\Delta t$ is taken to be 0.01s and control input $u$ is sampled randomly.  The training hyper-parameters used for training the DRKN are shown in Table~\ref{tab:nn_params}.

\begin{table}[ht!]
\small\centering
\caption{\label{tab:nn_params} Hyperparameters of the Koopman network along with the dataset information ( (T,S) : T - number of trajectories, S - Number of snapshots in each trajectory)}
    \begin{tabular}{ |p{1.5 cm}|p{1.75 cm}|p{1.75cm}|p{1.75 cm}| } 
     \hline
      & {Van der Pol Oscillator} & {4R manipulator} & {Franka FR3 manipulator} \\ 
     \hline
     \multicolumn{4}{|c|} {DRKN Autoencoder}\\
     \hline
    {Architecture} & [2, 20, 20, 20, 10] & [8, 40, 40, 40, 20] & [14, 30, 30, 30, 60] \\ \hline
    {\ Lifted state} & 12 & 28 & 74 \\\hline
     {$\alpha_1$}, {$\alpha_2$}, {$\alpha_3$} & 1, 0.5, 0.01 & 1, 0.5, 0.01 & 1, 0.5, 0.01\\  \hline
     {$\gamma_1$}, {$\gamma_2$} & 0, 0 & 0, 0 & 0, 0\\ \hline
     {Batch Size} & 256 & 256 & 256\\
     \hline
     {Optimizer} & adam & adam & adam \\ 
     \hline
     {Dataset} & (100, 100) & (350, 350) &  (820, 1000) \\
       
     \hline
    \end{tabular}
\end{table}

\begin{table}[ht!]
\small\centering
\caption{\label{tab:vand_pol_error_comp} Mean prediction error comparison for the Van der Pol oscillator.}
    \begin{tabular}{ |C{1.30cm}|C{1.30cm}|C{1.30cm}|C{1.30cm}|C{1.30cm}| } 
    \hline
    \multicolumn{5}{|c|} {Van der Pol Oscillator}\\
    \hline
     {Noise} & {Proposed} & {DKND \cite{Hao2024DeepKL}} & {Recurrent Rollout \cite{Sakib2024LearningNS}} & {NLK \cite{singh2024adaptive}}\\ 
     \hline
     {20dB} & {\textbf{0.097}} & {0.476} & {0.127} & {0.466} \\
     \hline
      {25dB} & {\textbf{0.077}} & {0.217} & {0.089} & {0.225} \\
     \hline
     {30dB} & {\textbf{0.068}} & {0.092} & {0.088} & {0.106} \\
     \hline
     {35dB} & {\textbf{0.061}} & {0.069} & {0.087} & {0.074} \\
     \hline
     {40dB} & {0.058} & {0.065} & {\textbf{0.054}} & {0.070} \\
     \hline
    \end{tabular}
\end{table}
The simulation experiments are conducted on Intel i7-12700 CPU.
The corresponding results for an open-loop prediction performance of DRKN is shown in Fig.\ref{fig:van_pol_sim}. The prediction performance is quantified by the average prediction error defined as $e_{pred} = (1/N_{snaps})\sum_{i=1}^{N_{snaps}} ||\boldsymbol{x_{i,true}} - \boldsymbol{x_{i,pred}} ||_2$ where $\boldsymbol{x_{true}}$ and $\boldsymbol{x_{pred}}$ are the true and predicted state trajectories, respectively, and $N_{snaps}$ is the number of time steps considered for the prediction. As illustrated in Fig.\ref{fig:van_pol_sim}b and Table \ref{tab:vand_pol_error_comp}, DRKN significantly outperforms both DKND \cite{Hao2024DeepKL} and NLK \cite{singh2024adaptive} frameworks in prediction accuracy. The NLK architecture does not account for the noise contained in the data, leading to an inaccurate prediction of the Koopman operator for increasing levels of noise in training data. The DKND\cite{Hao2024DeepKL} method augments its loss function with an additional term that penalizes discrepancies between the Koopman operators learned from noisy and noise-free data. However, as the true Koopman operator remains unknown, this formulation substitutes the error with an overly conservative upper bound. Consequently, the optimizer primarily minimizes this conservative bound by adjusting the basis functions, while largely neglecting the dominant influence of noise itself. As a result, DKND is relatively ineffective in mitigating the intrinsic bias introduced by training noise in (\ref{eq:Koopman_fwd_noise}), which quantifies how noise propagates in the learned Koopman operator for a given set of basis functions. In contrast, the proposed DRKN approach (\ref{eq:ReducedBiasKoop}) cancels this significant part of the noise bias and enables more accurate learning of the Koopman operator from noisy data. The approach in~\cite{Sakib2024LearningNS} utilizes polyflow basis functions and progressively increases the prediction horizon when computing the prediction loss during training. While it attains prediction performance that is comparable to DRKN for the Van der Pol oscillator, this approach incurs substantially higher computational costs due to the progressively increasing rollout length required to compute the prediction loss, as shown in Fig.~\ref{fig:van_pol_sim}c and Table~\ref{tab:vand_pol_error_comp}. As such, DRKN provides a favorable trade-off between accuracy and computational efficiency, making it a more practical choice for model learning in real-world scenarios.

\begin{figure}[ht!]
    \centering
    \includegraphics[width=\linewidth]
     {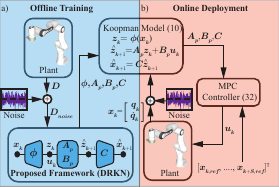}
     \caption{Schematic of the proposed Koopman-MPC framework} 
     \label{fig:MPC_schematics}
\end{figure}
\subsection{Model predictive control}
For the tracking control studies that follow, the proposed architecture is integrated with a linear Model Predictive Control (MPC) scheme as shown in Fig.~\ref{fig:MPC_schematics}, which leverages the learnt Koopman model to achieve efficient and reliable performance under joint/input constraints. The linear MPC essentially solves the following optimization problem for the prediction horizon H, 
\begin{align}
\label{mpc_framework}
 \min_{\boldsymbol{Z,U}}\sum^{H-1}_{k=0}\begin{bmatrix}
                \boldsymbol{Cz_{k} - x_{ref,k}}\\
                \boldsymbol{u_{k}}\end{bmatrix}^{T}\boldsymbol{Q}\begin{bmatrix}
                \boldsymbol{Cz_{k} - x_{ref,k}}\\
                \boldsymbol{u_k}
    \end{bmatrix} , 
\end{align}
s.t $\boldsymbol{z_{k+1}} = \boldsymbol{A_p}\boldsymbol{z_{k}} + \boldsymbol{B_p}\boldsymbol{u_{k}}, \boldsymbol{z_0} = \boldsymbol{\phi(x_0)}, \boldsymbol{x^{-}} \leq \boldsymbol{Cz_k}\leq \boldsymbol{x^{+}},\\ 
\boldsymbol{u^{-}} \leq \boldsymbol{u_k}\leq \boldsymbol{u^{+}}, k = 0,1, ..., H-1 $
where $\boldsymbol{x_{ref,k}}$ refers to the reference state at $k^{th}$ timestep of the prediction horizon. $\boldsymbol{Q}$ represents penalty matrix combined for both states and inputs, while state and input bound constraints are represented by $[\boldsymbol{x^{-},x^{+}}]$ and $[\boldsymbol{u^{-},u^{+}}]$ respectively. 
\subsection{4R manipulator simulation results}
Next, we evaluate our proposed architecture on a 4R serial manipulator operating within a three-dimensional workspace and implemented using Python Robotics Toolbox \cite{9561366}. For the 4R manipulator, the model parameters used for data collection are as follows: for the $i^{\text{th}}$ link, mass is $m_i = 0.1$~kg, length is $l_i = 0.33$~m, center of mass is located at $a_i = l_i/2$, and moment of inertia is $I_i = \text{diag}[0,1.5,1.5]$~kg$\cdot$m$^2$. We benchmark the proposed method against relevant state-of-the-art approaches in terms of both prediction accuracy and tracking performance, where the latter is evaluated under the MPC framework \eqref{mpc_framework} tasked with tracking sinusoidal joint-space trajectories. The corresponding hyperparameter and dataset information is presented in Table~{\ref{tab:nn_params}}.

\begin{figure}[ht!]
    \centering
    \includegraphics[width=\linewidth]
     {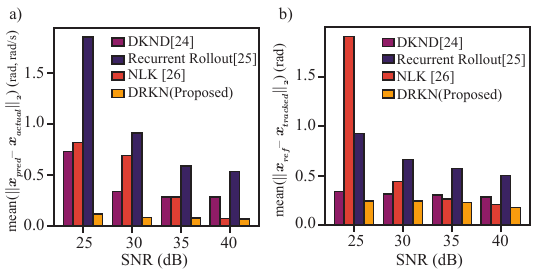}
    
     \caption{ a) Prediction performance comparison for learning dynamics of the 4R manipulator arm. b) Closed-loop tracking error comparison with feedback corrupted by 30 dB noise.} 
     \label{fig:4R_comp_all_pred_tracking}
\end{figure}
\begin{figure}[ht!]
    \centering
    \includegraphics[width=\linewidth]
     {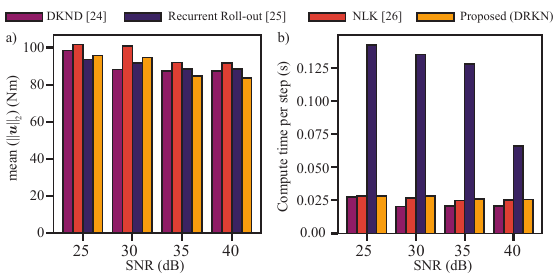}
    
     \caption{ Closed-loop comparison of a) control effort, and b) per-step MPC computation time for the 4R manipulator using Koopman models trained at different noise levels with feedback measurements corrupted by 30dB noise.} 
     \label{fig:4R_comp_all_control_effort}
\end{figure}

The training datasets for the 4R system are first corrupted with noise corresponding to SNR values between 25$-$40 dB. These noisy datasets are then used to learn the Koopman operators for all approaches under comparison, including the proposed (DRKN) method, DKND~\cite{Hao2024DeepKL}, Recurrent Roll-out~\cite{Sakib2024LearningNS}, and the NLK model~\cite{singh2024adaptive}. The corresponding open-loop prediction errors are computed 
against the true trajectory, with the respective errors shown in Fig. \ref{fig:4R_comp_all_pred_tracking}. Clearly, DRKN again outperforms DKND \cite{Hao2024DeepKL}, Recurrent Roll-out \cite{Sakib2024LearningNS}, and NLK \cite{singh2024adaptive} architectures in terms of prediction performance, demonstrating its ability to accurately capture the nonlinear system dynamics while maintaining robustness to measurement noise in the training data.
Subsequently, to evaluate tracking control performance, the previously learned Koopman models are now integrated within the Model Predictive Control (MPC) framework, tasked with tracking sinusoidal joint-space trajectories as shown in Fig. \ref{fig:MPC_schematics}. While tracking, the feedback measurements are further corrupted with significant gaussian noise corresponding to SNR of 30dB.   Tracking performance is quantified by mean tracking error defined as $e_{track} = (1/N)\sum_{i=1}^{N} ||\boldsymbol{\theta_{i,ref}} - \boldsymbol{\theta_{i,track}} ||_2$ where $\boldsymbol{\theta_{ref}}$ and $\boldsymbol{\theta_{track}}$ are reference and tracked joint angles, respectively, and $N$ is the number of time steps considered for tracking. It is important to emphasize that noise is introduced into the feedback path solely to emulate realistic operating conditions during closed-loop tracking. While this is useful to demonstrate the efficacy of the learnt model in closed-loop control, the development of accurate models from noisy data is the primary contribution of this study.

Figure \ref{fig:4R_comp_all_pred_tracking} presents the tracking errors in joint positions corresponding to the Koopman models learned from training datasets for the various noise levels. A quantitative comparison of the prediction and tracking error metrics are shown in Table~\ref{tab:4R_error_comp}.  Clearly, the proposed DRKN method consistently outperforms the DKND \cite{Hao2024DeepKL}, Recurrent Roll-out \cite{Sakib2024LearningNS} and NLK \cite{singh2024adaptive} baselines, exhibiting improved tracking performance aligned with its superior prediction accuracy, while requiring comparable control-effort and compute time as presented in Fig. \ref{fig:4R_comp_all_control_effort}. This improvement can be attributed to DRKN’s ability to learn more accurate system dynamics even in the presence of noise, which directly enhances the efficacy of the MPC controller within the closed-loop. 
\begin{table}[ht!]
\small\centering
\caption{\label{tab:4R_error_comp}  Mean prediction and tracking errors comparisons for 4R manipulator for different levels of noise in training datasets. Prediction error is computed against noise-less trajectories. For closed-loop tracking studies, the feedback state measurement is corrupted by 30~dB noise.\color{black}}
    \begin{tabular}{ |C{1.30cm}|C{1.30cm}|C{1.30cm}|C{1.30cm}|C{1.30cm}| } 
    \hline
     \multicolumn{5}{|c|} {4R Manipulator -- Prediction error (rad,\,rad/s)}\\
     \hline
     {Noise} & {Proposed} & {DKND \cite{Hao2024DeepKL}} & {Recurrent Rollout \cite{Sakib2024LearningNS}} & {NLK \cite{singh2024adaptive}}\\
     \hline
     {25dB} & {\textbf{0.114}} & {0.731} & {1.853} & {0.816} \\
     \hline
     {30dB} & {\textbf{0.081}} & {0.337} & {0.914} & {0.689} \\
     \hline
     {35dB} & {\textbf{0.074}} & {0.281} & {0.590} & {0.282} \\
     \hline
     {40dB} & {\textbf{0.063}} & {0.279} & {0.533} & {0.070} \\
     \hline
     \multicolumn{5}{|c|} {4R Manipulator -- Tracking error (rad)}\\
     \hline
     {25dB} & {\textbf{0.245}} & {0.340} & {0.923} & {1.904} \\
     \hline
     {30dB} & {\textbf{0.241}} & {0.316} & {0.663} & {0.442} \\
     \hline
     {35dB} & {\textbf{0.227}} & {0.306} & {0.570} & {0.266} \\
     \hline
     {40dB} & {\textbf{0.178}} & {0.282} & {0.502} & {0.209} \\
     \hline
     \multicolumn{5}{|c|} {4R Manipulator -- Control effort (Nm)}\\
     \hline
     {25dB} & {95.62} & {98.42} & {\textbf{93.60}} & {101.54} \\
     \hline
     {30dB} & {94.69} & {\textbf{88.18}} & {91.83} & {100.84} \\
     \hline
     {35dB} & {\textbf{84.75}} & {87.47} & {88.40} & {91.94} \\
     \hline
     {40dB} & {\textbf{83.60}} & {87.34} & {88.39} & {91.56} \\
     \hline
     \multicolumn{5}{|c|} {4R Manipulator -- Compute time per MPC iteration (sec)}\\
     \hline
     {25dB} & {0.0283} & {\textbf{0.0277}} & {0.142} & {0.0281} \\
     \hline
     {30dB} & {0.0281} & {\textbf{0.0204}} & {0.135} & {0.0267} \\
     \hline
     {35dB} & {0.0258} & {\textbf{0.0208}} & {0.127} & {0.0250} \\
     \hline
     {40dB} & {0.0256} & {\textbf{0.0205}} & {0.066} & {0.0253} \\
     \hline
     
    \end{tabular}
\end{table}

\subsection{Franka FR3 simulation results}

\begin{figure*}[h] 
\centering
    \includegraphics[width=\linewidth]{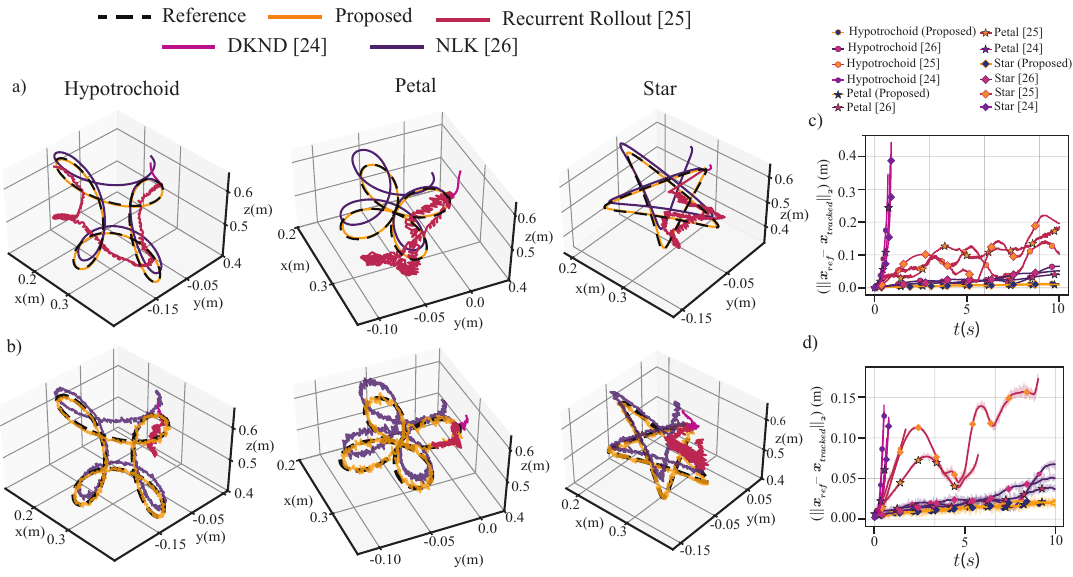}
    \caption{Trajectory tracking on the Franka FR3 manipulator using different schemes with Koopman models trained on noisy data (SNR = 30~dB). Tracking performance under (a) low ($\sim$80~dB) and (b) high (40~dB) measurement noise is shown. The corresponding tracking error evolution is reported for (c) low ($\sim$80~dB) and (d) high (40~dB) measurement noise.}
    \label{fig:franka_gaz_tracking}
\end{figure*}
Next, we assess the performance of the proposed algorithm on a 7-DoF Franka FR3 robotic arm simulated in the high fidelity Gazebo environment. The dataset employed for learning consists of 820 randomly generated trajectories, each consisting of 1000 snapshots sampled at 200 Hz. To evaluate the noise robustness of the proposed architecture, zero-mean Gaussian noise with a signal-to-noise ratio (SNR) of 30 dB is injected into both states (joint positions and velocities) and torque input measurements. 
The proposed method is again evaluated against the NLK \cite{singh2024adaptive}, DKND \cite{Hao2024DeepKL} and Recurrent Roll-out \cite{Sakib2024LearningNS} architectures in terms of trajectory tracking performance within the MPC framework \eqref{mpc_framework} to command Franka FR3 using torque control.  The evaluation focuses on the end-effector of a robotic arm tasked with following various target shapes. As illustrated in Fig.~\ref{fig:franka_gaz_tracking}a and Table \ref{tab:franka_fr3_error_comp}, the proposed architecture, trained on noisy data, demonstrates superior tracking accuracy compared to the Koopman models obtained from other methods. In particular, the approach in \cite{Sakib2024LearningNS}, which employs the polyflow basis functions for learning the Koopman model, shows poor tracking, while the MPC fails to converge with the DKND model after certain iterations. 

While the previous comparisons clearly demonstrate the robustness of the proposed method to noise in the training data, additional studies are undertaken that simulate more realistic conditions with increased levels of measurement noise. Specifically, measurement noise with SNR 40 dB is added to the feedback signals of joint positions and velocities used to close the control loop. Note that these noise levels are very similar to that expected with hardware implementation. In this case, the MPC fails to converge for the methods in \cite{Sakib2024LearningNS,Hao2024DeepKL} after some iterations. As illustrated in Fig.~\ref{fig:franka_gaz_tracking}b and Table \ref{tab:franka_fr3_error_comp}, the end-effector continues to follow the target shapes with the proposed (DRKN) approach, albeit with a slight degradation in tracking performance, which is expected due to higher noise levels in the feedback measurements. Thus, the proposed DRKN framework clearly delivers robust and reliable control performance under noisy conditions, outperforming existing approaches in both prediction and closed-loop tracking tasks.

\begin{figure*}[h] 
\centering
    \includegraphics[width=\linewidth]{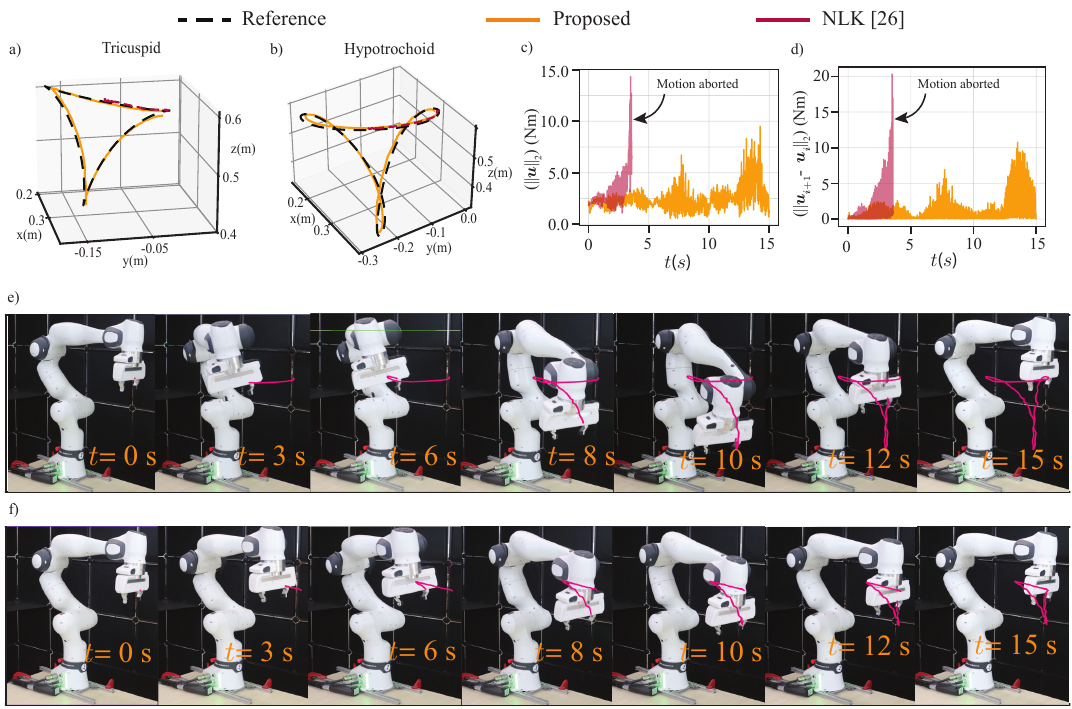}
    \caption{Real-world trajectory tracking with the Franka FR3 manipulator arm using Koopman models learned from data corrupted with 30 dB SNR noise for tracking (a,f) tricuspid and (b,e) hypotrochoid path. (c), (d) Control torque and torque rate profiles generated by the MPC-GESO framework for the Adobe trajectory using the proposed (DRKN) and NLK \cite{singh2024adaptive} Koopman models. The other architectures (DKND \cite{Hao2024DeepKL} and Recurrent-Rollout \cite{Sakib2024LearningNS}) are not shown as they induce instability immediately.}

    \label{fig:franka_real_tracking}
\end{figure*}
\begin{table*}
    \small \centering
    \caption{\label{tab:franka_fr3_error_comp}Performance comparisons (Mean tracking error (m), control effort (N-m) and compute-time-per-iteration (sec)) comparisons for the various schemes trained on the Franka FR3 robotic arm using simulated datasets with SNR 30dB.}
    \begin{tabular}{|C{1.30cm}|C{1.30cm}|C{1.30cm}|C{1.30cm}|C{1.30cm}|C{1.30cm}|C{1.30cm}|C{1.30cm}|C{1.30cm}|}
    \hline
    \multicolumn{1}{|c|}{} &
     \multicolumn{3}{c|} { \makecell[c]{Gazebo Simulation \\Case 1: Low feedback noise (80 dB)}}&
     \multicolumn{3}{c|}{\makecell[c]{Gazebo Simulation \\Case 2: High feedback noise (40 dB)}} &
     \multicolumn{2}{c|}{\makecell[c]{Real world experiment\\ High feedback noise - 40dB}} \\
     \hline
     Method&Hypo-trochoid & Petal & Star & Hypo-trochoid & Petal & Star & Hypo-trochoid & Tricuspid\\
     \hline
      \cite{Hao2024DeepKL} & - & - & - & - & - & - & - & - \\
     \hline
     \cite{Sakib2024LearningNS} & \makecell[c]{0.074\\(22.12)\\(0.0046)} & \makecell[c]{0.0963\\(19.01)\\(0.0045)} & \makecell[c]{0.0962\\(11.57)\\(0.0038)} & - & - & - & - & - \\
     \hline
      \cite{singh2024adaptive} & \makecell[c]{0.0231\\(\textbf{0.94})\\(0.0031)} & \makecell[c]{0.0188\\(\textbf{0.60})\\(0.0034)} & \makecell[c]{0.0254\\(\textbf{0.66})\\ \textbf{(0.0029)}} & \makecell[c]{0.023\\(\textbf{1.03})\\ \textbf{(0.0035)}} & \makecell[c]{0.021\\(\textbf{0.70})\\ \textbf{(0.0033)}} & \makecell[c]{0.025\\(\textbf{0.78})\\ \textbf{(0.0034)}} & - & - \\
     \hline
     Proposed& \makecell[c]{\textbf{0.0075} \\ (1.05)\\ \textbf{(0.0031)}} & \makecell[c]{\textbf{0.0063}\\ (0.65)\\ \textbf{(0.0030)}} & \makecell[c]{\textbf{0.0064}\\(0.70)\\ (0.0030)} & \makecell[c]{\textbf{0.012}\\(1.44)\\ (0.0042)} & \makecell[c]{\textbf{0.013}\\(1.12)\\ (0.0042)} & \makecell[c]{\textbf{0.014}\\(1.21)\\ (0.0043)} & \makecell[c]{\textbf{0.014}\\(\textbf{2.43})\\(\textbf{0.0099})} & \makecell[c]{\textbf{0.011}\\(\textbf{2.32})\\(\textbf{0.0098})} \\
     \hline
    \end{tabular}
\end{table*}
\section{Experimental Results}

This section evaluates the tracking performance of the proposed method for the Franka FR3 robotic arm in a real world environment. The same Koopman model, trained on Gazebo simulation data described in the previous section, is now deployed within the MPC framework \eqref{mpc_framework}. In order to mitigate the significant sim-to-real gap caused by training with simulated datasets, the Generalized Extended State Observer (GESO) \cite{singh2025generalizedmomentabasedkoopmanformalism} is now integrated with the MPC framework for robust disturbance rejection associated with unmodeled dynamics. It should be noted that all the Koopman models considered for comparisons are integrated within the MPC-GESO framework to ensure a fair basis for comparison. As shown in Fig. \ref{fig:franka_real_tracking}, the proposed scheme successfully tracks the desired trajectory with low tracking error for both target shapes considered in the experiments. As shown in Table \ref{tab:franka_fr3_error_comp}, the NLK \cite{singh2024adaptive}, DKND \cite{Hao2024DeepKL} and Recurrent-Rollout \cite{Sakib2024LearningNS} architectures fail to complete the task. The MPC-GESO controller fails to converge for the methods in \cite{Sakib2024LearningNS,Hao2024DeepKL}, owing to a very large sim-to-real gap of the corresponding Koopman models. Moreover, measurement noise of SNR 30 dB induces a very significant bias in model \cite{singh2024adaptive}, causing the MPC-GESO controller to produce incompatible torque and jerk (torque derivative) commands (Fig. \ref{fig:franka_real_tracking}d). The elevated control effort and sharp torque rate variations observed in Fig.~\ref{fig:franka_real_tracking}c-d are a direct consequence of the poor  Koopman model learned under noisy training conditions. These high-derivative torque inputs induce significant jerk in the Franka robotic arm, ultimately exceeding admissible thresholds and prompting the Franka control interface to abort the task. Similar trends are observed when the Koopman model is trained on data corrupted with 40dB noise, however these results are not included here for the sake of brevity. Thus, this experimental evaluation on the Franka FR3 manipulator in a real-world setting highlights the importance of learning a Koopman model directly from noisy data, particularly when the system is subject to practical constraints such as torque-command feasibility. The superior tracking performance and successful task execution demonstrate the effectiveness of the proposed framework in capturing an accurate Koopman representation under noisy conditions, thereby enhancing its suitability for real-time control.

\section{Conclusion}

This study proposes a Deep Robust Koopman Network (DRKN) to learn accurate Koopman models from noisy data. The proposed architecture simultaneously learns the forward and backward temporal dynamics to mitigate the bias introduced by noise contained in training data. A detailed theoretical analysis is used to provide robust predictive performance guarantees for the Koopman operator obtained using the proposed scheme. Simulation studies of open loop predictive performance and closed loop trajectory tracking are used to demonstrate the efficacy of the proposed framework across multiple manipulator systems. Detailed performance comparisons are further used to demonstrate the efficacy of the proposed scheme in learning the noise-robust Koopman operator relative to leading alternative designs. Finally, experimental validation studies in a real-world setting clearly highlight the importance of synthesizing noise-robust Koopman models of dynamical systems operating under realistic constraints. Future work would involve extending the proposed scheme by relying on real-world data to update the Koopman operator online for improved task-specific generalizability. 
\bibliographystyle{ieeetr}
\bibliography{citation}

\end{document}